\documentclass{article}
\usepackage{spconf,amsmath,graphicx,bm,amsfonts, multirow, float, array,ctable}

\title{MetaGrad: Adaptive Gradient Quantization with Hypernetworks}

\name{Kaixin Xu$^{1,4}$, Alina Hui Xiu Lee$^{3}$, Ziyuan Zhao$^{1,2}$, Zhe Wang$^{1,2,4}$, Min Wu$^{1,2}$, Weisi Lin$^{4}$}
\address{$^{1}$Institute for Infocomm Research, A*STAR, Singapore\\
$^{2}$ Artificial Intelligence, Analytics And Informatics (AI$^3$), A*STAR, Singapore\\
$^{3}$ National University of Singapore\\
$^{4}$School of Computer Science and Engineering, Nanyang Technological University, Singapore}
\begin{document}
%
\maketitle
\begin{abstract}
A popular track of network compression approach is Quantization aware Training (QAT), which accelerates the forward pass during the neural network training and inference. However, not much prior efforts have been made to quantize and accelerate the backward pass during training, even though that contributes around half of the training time. This can be partly attributed to the fact that errors of low-precision gradients during backward cannot be amortized by the training objective as in the QAT setting. In this work, we propose to solve this problem by incorporating the gradients into the computation graph of the next training iteration via a hypernetwork. Various experiments on CIFAR-10 dataset with different CNN network architectures demonstrate that our hypernetwork-based approach can effectively reduce the negative effect of gradient quantization noise and successfully quantizes the gradients to INT4 with only $0.64$ accuracy drop for VGG-16 on CIFAR-10.
\end{abstract}
\section{Introduction}
\label{sec:intro}
Model compression~\cite{zhou2016dorefa,Jin_2020_CVPR,zhe2019optimizing,Xu_2023_ICCV} is an important technique to reduce the model size and computation complexity of deep models. Among which, quantization-aware Training (QAT) is a popular track of research that simulates the neural network quantization (weights and activations) during the course of training to curb the inference-time accuracy drop of low-bit models (\emph{e.g.} INT8 quantization). 
On the other hand, theoretical and empirical analysis~\cite{jcjohnson_2016,banner2018scalable} show that backpropagation accounts for more computations than forward during training. 
By quantizing the gradients in the backward pass, the training of QAT models can be further accelerated utilizing the low-precision compute kernels. 

Despite this huge potential benefit, quantizing backward gradients is particularly difficult compared to quantizing forward variables, as backward variables are not included in the computation graph and thus cannot be optimized by the objective function, which could diverge the parameter update towards the wrong directions. 
In the past, a few gradient quantization works have been proposed~\cite{zhu2020towards,zhao2021distribution} that primarily achieved stable INT8 forward and backward quantization by designing gradient quantizers tailored to gradient characteristics. However, due to the abovementioned difficulties, lower than INT8 quantization is extremely under-explored, with the only early attempt DoReFa~\cite{zhou2016dorefa} ending up with poor results under INT6 gradients. This shows that designing a gradient quantizer with heuristics has already pushed to its limit. 

To address the challenge of low-bit gradient quantization and curb its great performance degradation, in this paper, we propose a novel quantization approach that is adaptive to arbitrary quantization noise and best maintains the stable training direction. This meta-quantizer is realized by incorporating the hypernetwork~\cite{ha2017hypernetworks}, which is trained end-to-end along with the network to predict the quantized version of the gradient from a given full-precision gradient. In this regard, the low-precision gradients are now loss-aware since we incorporate them into the computation graph, resulting in a more robust QAT training. The proposed meta gradient quantizer generates higher test performance on various CNNs architectures.

\section{Related Works}

\subsection{Quantization-aware Training (QAT)}
DoReFa-Net~\cite{zhou2016dorefa} proposed to optimize the clipping value and the scaling factor of the uniform quantizers for weights and activations separately. It was validated on image classification tasks under multiple bit-widths, but only with the rather simple AlexNet architecture. 
Most QAT works quantize weights and activations simultaneously by optimizing the uniform quantization parameters~\cite{zhang2018lq,esser2019learned,bhalgat2020lsq}, layer-wise or channel-wise mixed-precision quantization~\cite{Jin_2020_CVPR,autoq}, or leveraging non-uniform quantization such as Logarithmic quantizer~\cite{miyashita2016convolutional}.
Most recent QAT works~\cite{zhou2016dorefa,esser2019learned,bhalgat2020lsq} used ``Straight-Through Estimator" (STE)~\cite{Bengio2013} to estimate the gradient of the non-differentiable quantization function, while another work~\cite{gong2019differentiable} softened the linear quantization operation in order to match the true gradient with STE.

\subsection{Gradient Quantization in QAT}
An earlier attempt~\cite{zhou2016dorefa} adopted a primitive quantizer design based on a uniform quantizer for gradients (without scaling and other optimization), and large performance drops were observed when training with low-bit gradients. 
SBM~\cite{banner2018scalable} adopted fixed-point 8-bit gradient quantization but only focused on improving the quantization schemes in the forward pass. WAGEUBN~\cite{yang2020training} quantized gradients to 8-bit integers but also showed a huge performance gap against its full-precision counterpart. In UINT8~\cite{zhu2020towards}, the authors considered the sharp and wide distribution of gradients and proposed to clip the gradients according to the deviation of the gradient distribution before quantization, achieving results on-par with full-precision training.
To compensate the quantization loss on gradient, AFP~\cite{zhang2020fixed} and CPT~\cite{fu2021cpt} used higher precision data to aid low-precision training. DAINT8~\cite{zhao2021distribution} adopted a bespoke 8-bit channel-wise gradient quantization to suppress the negative effect of quantization noise during training.
Gradient quantization with less than INT8 representations remains largely unexplored. FP4~\cite{sun2020ultra} managed to train modern CNN architectures using 4-bit gradients without significant accuracy loss, but the gradients were represented as floating-point numbers.

\subsection{Hypernetworks}
Another interesting line of work 
uses the concept of hypernetwork, that is first proposed in~\cite{ha2017hypernetworks}, to turn non-differential parts in neural networks into differentiable operations. MetaPruning \cite{liu2019metapruning} trains a hypernetwork to generate sparse layer weights for structured pruning of CNNs automatically. Meta-Quant~\cite{chen2019metaquant} replaces
the STE~\cite{Bengio2013} in the QAT training with a hypernetwork to reduce the estimation error of STE. 

\section{Preliminaries}
Quantization is a process that maps full-precision (FP32) values to a set of discrete values. 
Under normal gradient quantization schemes~\cite{zhou2016dorefa,zhu2020towards,zhao2021distribution}, both error signals $\nabla_\mathbf{x} L$ w.r.t. activation $x$ calculated real-time during backpropagation and the gradient of weight $\nabla_\mathbf{W} L$ are in low-precision, which is similar to quantization scheme of forward pass. For visual simplicity, we omit the training loss $L$ in the latter text and express gradients as $\nabla_{(\cdot)}$.
We discuss integer quantization in this work. For typical gradient quantization (e.g., in~\cite{zhu2020towards}), symmetric uniform quantization is used. Given the gradients $\nabla L$ and a clipping value $c\in (0, \max{(\left|\nabla L\right|)}]$, the symmetric uniform quantizer with bit-width $B$ can be formulated as: 
\begin{equation}\label{eq:quantizer}
    (\nabla L)_q = Q(\nabla L)= \mathrm{round}\left(\mathrm{clip}(\nabla L,c)\cdot\frac{2^{B-1}-1}{c}\right),
\end{equation}
where $\mathrm{clip}(\nabla L, c)=\min(\max(\nabla L, -c), c)$. The quantized gradients are de-quantized as $\Tilde{\nabla L} = (\nabla L)_q \cdot\frac{c}{2^{B-1}-1}$.

\section{Methodologies}
\vspace{-10pt}
\label{sec:grad}
\begin{figure}[h!]
    \centering
    \includegraphics[width=.88\linewidth]{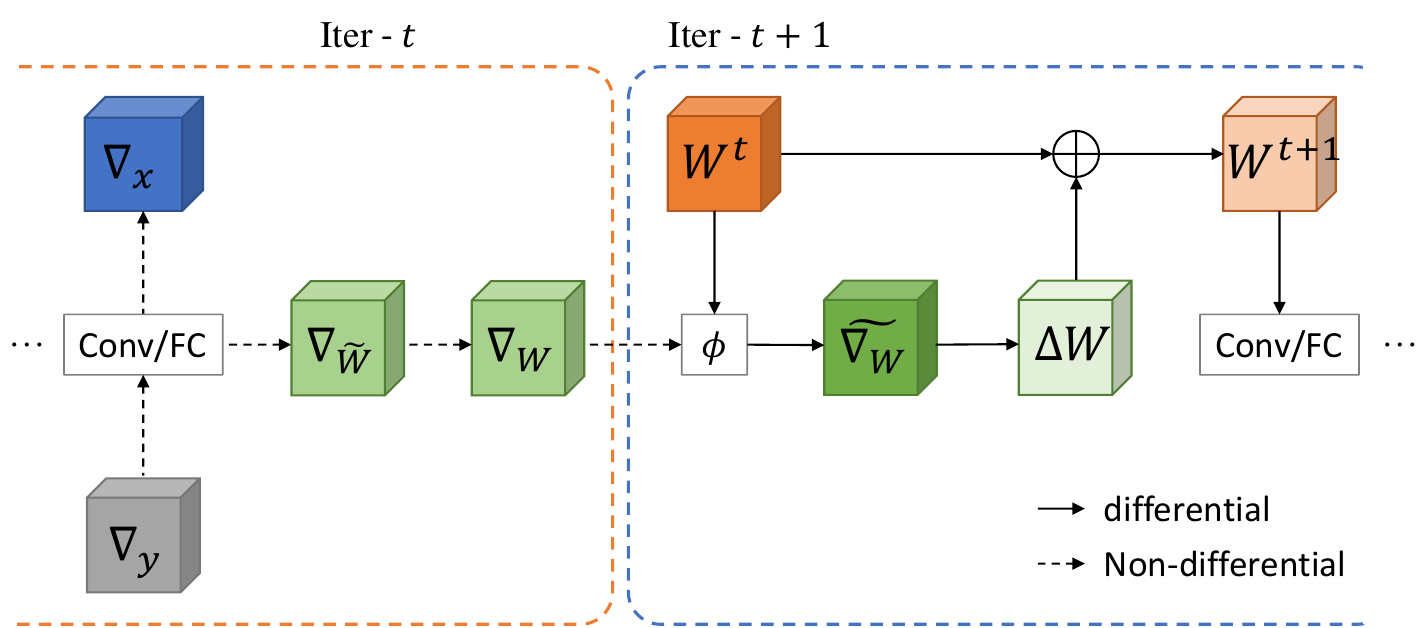}
    \caption{Illustration of delayed updating strategy of meta-quantization of gradients.}
    \label{fig:illu}\vspace{-10pt}
\end{figure}

\subsection{Meta-quantizer for Gradients}
\label{sec:metaq}
We now explain in detail the design of meta-quantizer for gradients using hypernetwork. 
The main challenge we aim to solve here is how to incorporate the quantization operation of gradients in backpropagation into a forward computation graph so that the gradient quantizer is 
loss-aware. To achieve this, we first define a light-weight, differentiable neural network $f_\phi$ that is shared across all quantizable layers in the main model $f$. $f_\phi$ is parameterized by $\phi$, and it inputs the propagated full-precision gradients or a combination of gradients and other variables (such as weight $\mathbf{W}$) and outputs the corresponding quantized gradient. The detailed structure of the hypernetwork $f_\phi$ will be discussed in latter sections.
Inspired by MetaQuant~\cite{chen2019metaquant}, we adopt the delayed updating strategy for layer weights (see Fig.~\ref{fig:illu}). Specifically, first, at the $t$-th iteration of training, when the backpropagation passing through a Conv/FC layer, $\nabla_\mathbf{W}$ is obtained by chain-rule, 
\vspace{-8pt}
\begin{equation}
\nabla_\mathbf{W} = \left[\frac{\partial \mathbf{y}}{\partial \mathbf{\Tilde{\mathbf{W}}}}\right]^\top \left[\frac{\partial \Tilde{\mathbf{W}}}{\partial \mathbf{W}}\right]^\top \nabla_\mathbf{y},
\vspace{-8pt}
\end{equation}
where $\Tilde{\mathbf{W}}$ is the quantized weight in forward pass. Next, we use meta-quantizer to quantize $\nabla_\mathbf{W}$ before updating $\mathbf{W}$. This is realized by treating full-precision $\nabla_\mathbf{W}$ as input to the hypernetwork $f_\phi$, which outputs the low-precision gradient $\widetilde{\nabla_\mathbf{W}}$, given by,
\vspace{-8pt}
\begin{equation}
    \widetilde{\nabla_\mathbf{W}} = f_\phi(\nabla_\mathbf{W}, \mathbf{W}).
\end{equation}
To obtain the weight update amount $\Delta \mathbf{W}$ for iteration $t+1$, the quantized gradient is then further refined by optimization strategy $\pi$ given by e.g. SGD with momentum or Adam method, and scaled by learning rate $\mu$, given by,
\vspace{-8pt}
\begin{equation}
    \Delta\mathbf{W} = -\mu\pi\left(\widetilde{\nabla_\mathbf{W}}\right),
\vspace{-8pt}
\end{equation}

where the implementation of $\pi$ varies according to different optimization options, which is typically differentiable for common optimizers such as SGD and Adam. Specifically, for SGD, $\pi\left(\widetilde{\nabla_\mathbf{W}}\right)=\widetilde{\nabla_\mathbf{W}}$.
From Fig.~\ref{fig:illu}, one can see that all the computations in between gradient $\nabla_\mathbf{W}$ and the weight update $\mathbf{W}^{t+1}$ are differentiable, and we successfully incorporate the meta-quantizer into the computation graph of the main network. 
Hence, the complete weight update procedure during the forward pass can be written as: 
\begin{equation}\label{eq:forw}
    \mathbf{W}^{t+1} = \mathbf{W}^t - \mu\pi\left(f_\phi(\nabla_\mathbf{W}, \mathbf{W})\right).
\end{equation}

\subsection{Design choices of hypernetwork}

\begin{figure}[h]
\vspace{-10pt}
    \centering
    \includegraphics[width=0.55\linewidth]{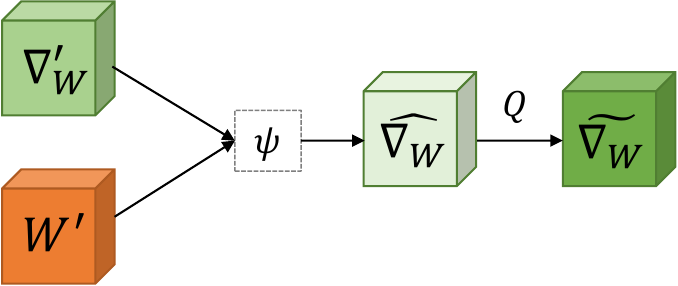}
    \caption{General architecture of the proposed hypernetwork for meta gradient quantization.}
    \label{fig:phi}\vspace{-10pt}
\end{figure}

As mentioned in Sec~\ref{sec:metaq}, in order to incorporate the gradients into the computation graph, the general design of hypernetwork $f_\phi$ should be a differentiable network that satisfies 
 \vspace{-8pt}
\begin{equation}
    f_\phi: \mathbb{R}^{k\times 1} \times \mathbb{R}^{k\times 1} \rightarrow \mathbf{S}^{k\times 1},
 \vspace{-8pt}
\end{equation}
where $\mathbf{S}$ is a finite set that satisfies
 \vspace{-8pt}
\begin{equation}
    \mathbf{S} = \left\{\frac{ck}{2^{B-1}-1} \mid -2^{B-1} \le k \le 2^{B-1}, k \in\mathbb{Z}\right\}.
 \vspace{-8pt}
\end{equation}
This can be achieved by letting $f_\phi=f_\psi\circ Q$, where $f_\psi: \mathbb{R}^{N\times 1} \times \mathbb{R}^{N\times 1} \rightarrow \mathbb{R}^{N\times 1}$ and $Q:\mathbb{R}^{N\times 1}\rightarrow\mathbf{S}^{N\times 1}$ is the quantizer defined in Eq.~\ref{eq:quantizer}. By assigning the gradient of $Q$ using STE, $f_\phi$ becomes differentiable. 

 With this, based on \cite{chen2019metaquant}, for FC layer, we can first transpose both the gradient $\nabla_\mathbf{W}\in\mathbb{R}^{n\times c}$ and the weight $\mathbf{W}\in\mathbb{R}^{(n)\times c}$ to $\mathbb{R}^{(nc)\times 1}$ tensors $\vec{\nabla_\mathbf{W}}$ and $\vec{\mathbf{W}}$ respectively, then input them to $f_\phi$; and for Conv layer kernels, we can transpose $\mathbb{R}^{n\times c\times k\times k}$ to $\mathbb{R}^{(nck^2)\times 1}$. 
 And for the design of $f_\psi$, we could simply adopts the two designs in \cite{chen2019metaquant}:
 \vspace{-8pt}
 \begin{equation}
     \text{\texttt{MultiFC:}} \quad f_\psi= \nabla_\mathbf{W}^\prime \cdot FCs(\mathbf{W}^\prime ),
     \vspace{-8pt}
 \end{equation}
  \vspace{-8pt}
 \begin{equation}
     \text{\texttt{LSTMFC:}} \quad f_\psi= \nabla_\mathbf{W}^\prime \cdot FCs(LSTM(\mathbf{W}^\prime )).
  \vspace{-8pt}
 \end{equation}
 However, we found that such coordinate-wise neural network design~\cite{andrychowicz2016learning} cannot make 
 good use of the intra-layer relations of the weights and gradients. Therefore, we design another $f_\psi$ as:
  \vspace{-8pt}
 \begin{equation}
     \text{\texttt{DualLSTMFC:}} \quad f_\psi= FCs(LSTM(\mathcal{J})),
 \vspace{-8pt}
 \end{equation}
where $\mathcal{J}=[\mathbf{W}^\prime;\nabla_\mathbf{W}^\prime]\in\mathbb{R}^{N\times 2}$ denoting a concatenation. This design can better exchange the information between weight and gradient tensors and potentially quantize the gradients in favor of the weight updating. 
In our empirical evaluations, we also observed a consistent improvement from \texttt{LSTMFC} design.

\subsection{Training of meta-quantizer}
According to Eq.~\ref{eq:forw}, the gradient received by the hypernetwork $f_\phi$ at $t+1$-th iteration can be obtained by chain-rule:
 \vspace{-8pt}
\begin{equation}
    \frac{\partial{L}}{\partial \phi^{t+1}} = \frac{\partial L}{\partial \mathbf{W}^{t+1}}\frac{\partial\mathbf{W}^{t+1}}{\partial\phi^{t+1}}=-\mu\pi^\prime \cdot \frac{\partial L}{\partial \mathbf{W}^{t+1}}\frac{\partial f_\phi(\nabla_\mathbf{W}, \mathbf{W})^\top}{\partial\phi^{t+1}}.
 \vspace{-8pt}
\end{equation}
We can further get the gradients w.r.t. $\psi$ according to $f_\phi=f_\psi\circ Q$ by: 
 \vspace{-8pt}
\begin{align}
    \frac{\partial{L}}{\partial \psi^{t+1}} &= \frac{\partial{L}}{\partial \phi^{t+1}}\frac{\partial \phi^{t+1}}{\partial \psi^{t+1}} \\\notag
    &\overset{\mathrm{STE}}{\approx} = \frac{\partial{L}}{\partial \phi^{t+1}} = -\mu\pi^\prime \frac{\partial L}{\partial \mathbf{W}^{t+1}}\frac{\partial f_\phi(\nabla_\mathbf{W}, \mathbf{W})^\top}{\partial\phi^{t+1}},
 \vspace{-8pt}
\end{align}
and therefore (omit superscript $t+1$):
 \vspace{-8pt}
\begin{equation}
    \nabla_\psi = \left[\frac{\partial{L}}{\partial \psi}\right]^\top \approx -\mu\pi^\prime \frac{\partial f_\phi(\nabla_\mathbf{W}, \mathbf{W})}{\partial\phi^{t+1}} \nabla_\mathbf{W}.
 \vspace{-8pt}
\end{equation}

Since the gradients of the meta-gradient-quantizer are defined, we can train the hypernetwork together with the main model in an end-to-end manner. 

\section{Experiments}
\subsection{Implementation Details}
We first tested different hypernetwork choices, namely MultiFC, LSTMFC and DualLSTMFC on 4 bit quantization of 2 architectures, ResNet-20 and VGG-16, using CIFAR-10 dataset. 
We conducted the experiments for multiple CNNs architectures on CIFAR-10 dataset for INT8 and INT4 QAT training. We compare ourselves to UINT8~\cite{zhu2020towards} and DAINT8~\cite{zhao2021distribution}. For fair comparison, we used dorefa~\cite{zhou2016dorefa} as the quantization method and \texttt{DualLSTMFC} as the hypernetwork choice.
For all experiments, SGD with momentum was chosen for optimizer.
For INT8 experiments, initial learning rate was 0.01 for ResNet-20 and 0.005 for ResNet-32, MobileNet-V2 and VGG-16, learning rate weight decay was 0.001 for ResNet-20 and 0.0001 for ResNet-32, MobileNet-V2 and VGG-16.
For INT4, learning rate weight decay of 0.0001 were used for all and for VGG-16 and MobileNet-V2, initial learning rate was 0.01 for ResNet-20 and 0.005 for ResNet-32, 0.001 for ResNet-56, 0.0005 for MobileNet-V2 and VGG-16 and 0.0001 for ResNet-110. 


\subsection{Experimental Results}

\begin{table}[!t]
\newcolumntype{C}[1]{>{\centering\arraybackslash}m{#1}}
\begin{center}
    \renewcommand{\arraystretch}{1.1}
    \setlength{\tabcolsep}{5pt}
    \begin{tabular}{c|c c c c} 
    \toprule[1.1pt]
        Model & Method & FP32 /\% &   Acc /\%
        & \(\Delta \) /\% \\ 
        \toprule[1.1pt]
        \multirow {2}{*}{ResNet-20}
        & UINT8 & 91.36 & 73.45 & -17.91\\
        & Ours & 91.36 & \textbf{91.2} & \textbf{-0.16}\\
        \midrule
        \multirow{2}{*}{ResNet-32}
        & UINT8 & 91.42 & 63.46 & -27.96 \\
        & Ours & 91.42 & \textbf{90.56} & \textbf{-0.86} \\
        \midrule
        \multirow{2}{*}{ResNet-56}
        & UINT8 & 91.95 & 65.27 & -26.68 \\
        & Ours & 91.95 & \textbf{90.24} & \textbf{-1.71} \\
        \midrule
        \multirow{2}{*}{ResNet-110}
        & UINT8 & 91.67 & 62.57 & -29.10 \\
        & Ours & 91.67 & \textbf{89.16}& \textbf{-2.51} \\
        \midrule
        \multirow{2}{*}{MobileNet-V2}
        & UINT8 & 93.57 & 88.12 & -5.45 \\
        & Ours & 93.57 & \textbf{89.00} & \textbf{-4.57} \\
        \midrule
        \multirow{2}{*}{VGG-16}
        & UINT8 & 92.25 & 22.23 & -70.02 \\
        & Ours & 92.25 & \textbf{91.61} & \textbf{-0.64} \\
        \toprule[1.1pt]
    \end{tabular}
    \caption{Comparison with other methods under INT4 quantization. $\Delta$ denotes the performance drop from full-precision (FP32) results: $\Delta=\text{Acc}-\text{FP}$ (the higher the better). }\vspace{-18pt}
    \label{tab:table2}
\end{center}
\end{table}
\noindent\textbf{INT4 Results.} 
Tab.~\ref{tab:table2} shows the comparison of Top-1 accuracies of different CNNs on CIFAR-10 INT4 quantization. 
For UINT8~\cite{zhu2020towards}, we report the results from our re-implementation using the same weight and activation quantizer schemes which is DoReFa. 
Under 4-bit, it is obvious that the proposed quantization scheme achieves much less accuracy degradation than UINT8 for all models, as shown in Tab.~\ref{tab:table2}.
Particularly for VGG-16, we even observe a whopping $69.38$ improvement from UINT8, which we suspect is due to the lack of residual structures in the VGG-16 model that exacerbates the negative effects of gradient quantization error. 
Our method consistently supports stabilized gradient quantization training, where the degradation from FP32 across all models is at most $4.57$.

\noindent\textbf{INT8 Results.} 
Tab.~\ref{tab:table1} shows the 
comparison of 
Top-1 accuracies of different CNNs on CIFAR-10 INT8 quantization.
We observe that under 8-bit, we perform the best compared to both state-of-the-art gradient quantization ~\cite{zhu2020towards,zhao2021distribution} under almost all examined models, especially for VGG-16, where the baseline UINT8 simply fail to converge, while our quantization scheme successfully maintains the performance drop with FP to only $-0.05$. The only exception is ResNet-20, where we perform slightly worse than DAINT8~\cite{zhao2021distribution}, but this is expected since DAINT8 uses the more fine-grained channel-wise quantization while ours and UINT8~\cite{zhu2020towards} use the de-facto simple layer-wise quantization. Yet, we still 
have a lower performance drop from FP32 than DAINT8 on MobileNet-V2, revealing the great potential of our approach that obtains performance gain only by incorporating the gradients into the objective function.

\subsection{Ablation Studies and Other Discussions}

\begin{table}[!t]
\newcolumntype{C}[1]{>{\centering\arraybackslash}m{#1}}
\begin{center}
    \renewcommand{\arraystretch}{1.1}
    \setlength{\tabcolsep}{5pt}
    \begin{tabular}{c|c c c c} 
    \toprule[1.1pt]
        Model & Method & FP32 /\% &   Acc /\%
        & \(\Delta \) /\% \\ 
        \toprule[1.1pt]
        \multirow {3}{*}{ResNet-20}
        & UINT8 &  91.36 & 91.10 & -0.26 \\
        & DAINT8 & 92.35 & \textbf{92.76} & \textbf{0.41} \\
        & Ours & 91.36 & 91.30 & -0.06\\
        \midrule
        \multirow{2}{*}{ResNet-32}
        & UINT8 & 91.42 & 90.70 & -0.72 \\
        & Ours & 91.42 & \textbf{91.40} & \textbf{-0.02} \\
        \midrule
        \multirow{3}{*}{MobileNet-V2}
        & UINT8 & 93.57 & 93.38 & -0.19 \\
        & DAINT8 & 94.73 & \textbf{94.37} & -0.36 \\
        & Ours & 93.57 & 93.51 & \textbf{-0.06} \\
        \midrule
        \multirow{2}{*}{VGG-16}
        & UINT8 & 92.25 & 31.32 & -60.93 \\
        & Ours & 92.25 & \textbf{92.20} & \textbf{-0.05} \\
        \toprule[1.1pt]
    \end{tabular}
    \caption{Comparison with other methods under INT8 quantization. }\vspace{-18pt}
    \label{tab:table1}
\end{center}
\end{table}

\begin{table}[!t]
\centering
  \begin{center}
    \begin{tabular}{c |c c c c}
    \toprule[1.1pt]
      Model & Hypernetwork & FP32 & Acc /\% & \(\Delta \) /\% \\
      \toprule[1.1pt]
      \multirow {3}{*}{ResNet-20} 
      & \texttt{MultiFC} & 91.36 & 91.06 & -0.30\\
      & \texttt{LSTMFC} & 91.36 & 91.15 & -0.21\\
      & \texttt{DualLSTMFC} &91.36 & \textbf{91.20} & \textbf{-0.16}\\
      \midrule
      \multirow {3}{*}{VGG-16}
      & \texttt{MultiFC} & 92.25 & 91.50 & -0.75\\
      & \texttt{LSTMFC} & 92.25 & 91.54 & -0.71\\
      & \texttt{DualLSTMFC} & 92.25 & \textbf{91.61} & \textbf{-0.64}\\
      \toprule[1.1pt]
    \end{tabular}
    \caption{Ablation study of hypernetwork design under INT4 quantization.}\vspace{-18pt}
    \label{tab:abl}
  \end{center}
\end{table}

We conduct an ablation study for the choice of hypernetwork design on CIFAR-10 with two models, ResNet-20 and VGG-16 under INT4 quantization. Tab.~\ref{tab:abl} shows a clear pattern of \texttt{DualLSTMFC} $>$ \texttt{LSTMFC} $>$ \texttt{MultiFC} for both models, indicating that our newly proposed \texttt{DualLSTMFC} design can more effectively generate quantized gradients in favor of the weight updating, resulting in the lowest accuracy degradation. Yet, the overall performance discrepancy among them is not large, indicating that the choice of hypernetwork only marginally affects the effectiveness of the proposed method.

\section{Conclusions}
In this work, to improve the stability of gradient quantization during QAT, we presented a novel meta-quantizer for gradient quantization that is able to automatically optimize the quantizer throughout training and thus more robust against quantization noise produced in the backward pass. Extensive experiments on the CIFAR-10 dataset with various CNNs demonstrate that the meta-quantizer helps the QAT training stabilize under lower gradient bit-rates, which vastly outperforms the baselines, especially under extremely low bit-widths.

\section*{Acknowledgement}
This research is supported by the Agency for Science, Technology and Research (A*STAR) under its Funds (Project Number A1892b0026, and C211118009). Any opinions, findings and conclusions or recommendations ex- pressed in this material are those of the author(s) and do not reflect the views of the A*STAR.

\newpage
\bibliographystyle{ieee_fullname}
\bibliography{ref}

\end{document}